\documentclass[10pt,twocolumn,letterpaper]{article}

\usepackage{cvpr}
\usepackage{times}
\usepackage{epsfig}
\usepackage{graphicx}
\usepackage{amsmath}
\usepackage{amssymb}

\usepackage{url}            
\usepackage{booktabs}       
\usepackage{nicefrac}       

\usepackage{subcaption}
\usepackage{bm}
\usepackage{xcolor}

\usepackage{enumitem}
\newcommand{\citet}{\cite}
\newcommand{\citep}{\cite}
\setlist{leftmargin=.3in}
\newcommand{\commentAD}[1]{}
\newcommand{\commentSA}[1]{}

\newcommand{\toappendix}[1]{}
\newcommand{\std}[1]{($\pm$ #1)}

\graphicspath{{.}}

\usepackage[breaklinks=true,bookmarks=false]{hyperref}

\cvprfinalcopy 

\begin{document}

\title{Variational Information Distillation for Knowledge Transfer}

\author{
Sungsoo Ahn \thanks{Contributed during an internship at Amazon.}\\
Korea Advanced Institute of Science and Technology\\
Daejeon, Korea\\
{\tt\small sungsoo.ahn@kaist.ac.kr}
\and
Shell Xu Hu $^{*}$\\
École des Ponts ParisTech \\
Champs-sur-Marne, France \\
{\tt\small hus@imagine.enpc.fr}
\and
Andreas Damianou \\
Amazon \\
Cambridge, United Kingdom \\
{\tt\small damianou@amazon.com}
\and
Neil D. Lawrence \\
Amazon \\
Cambridge, United Kingdom \\
{\tt\small lawrennd@amazon.com}
\and
Zhenwen Dai \\
Amazon \\
Cambridge, United Kingdom \\
{\tt\small zhenwend@amazon.com}
}

\maketitle

\begin{abstract}
Transferring knowledge from a teacher neural network pretrained on the same or a similar task to a student neural network can significantly improve the performance of the student neural network. Existing knowledge transfer approaches match the activations or the corresponding hand-crafted features of the teacher and the student networks. We propose an information-theoretic framework for knowledge transfer which formulates knowledge transfer as maximizing the mutual information between the teacher and the student networks. We compare our method with existing knowledge transfer methods on both knowledge distillation and transfer learning tasks and show that our method consistently outperforms existing methods. We further demonstrate the strength of our method on knowledge transfer across heterogeneous network architectures by transferring knowledge from a convolutional neural network (CNN) to a multi-layer perceptron (MLP) on CIFAR-10. The resulting MLP significantly outperforms the-state-of-the-art methods and it achieves similar performance to the CNN with a single convolutional layer.
\end{abstract}

\section{Introduction}





Deep neural networks (DNNs) play important roles in various computer vision tasks, \eg, depth estimation \cite{eigen2014depth}, pose estimation \cite{toshev2014deeppose}, optical flow \cite{dosovitskiy2015flownet}, object classification \cite{he2016deep}, detection \cite{girshick2015fast}, and segmentation \cite{simonyan2014very}. A typical DNN approach for a computer vision task is to train a sophisticated end-to-end neural network with a large amount of labeled data. Such an approach often delivers state-of-the-art performance if a sufficient amount of data is available. However, in many cases, it is impossible to gather sufficiently large data to train a DNN. For example, in many medical image applications \cite{schlegl2014unsupervised}, the amount of available data is constrained by the number of patients of a particular disease.

\begin{figure}
  \centering
  \includegraphics[width=.45\textwidth]{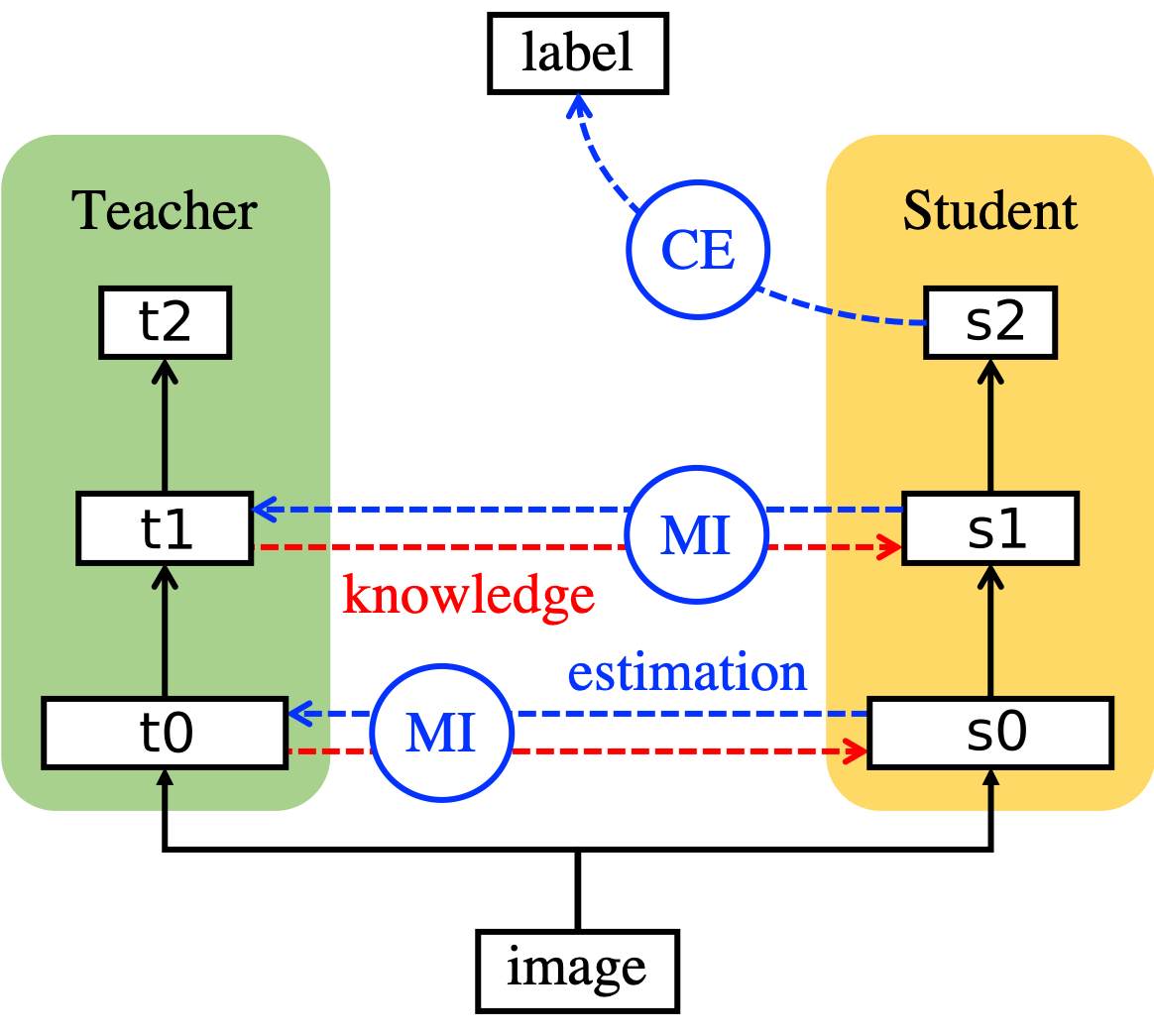}
\caption{Conceptual diagram of the proposed knowledge transfer method. The student network efficiently learns the target task by minimizing the cross-entropy (CE) loss while retaining high mutual information (MI) with the teacher network. The mutual information is maximized by learning to estimate the distribution of the activations in the teacher network, provoking the transfer of knowledge.}
\label{fig:diagram}
\end{figure}

A popular approach for handling such lack of data is transfer learning \cite{pan2010survey}, where the goal is to transfer knowledge from the source task to facilitate learning on the target task. Typically, one considers the source task to be generic with a larger amount of available data that contains useful knowledge for learning the target task, \eg, knowledge from natural image classification \cite{russakovsky2015imagenet} is likely to be useful for fine-grained bird classification \cite{WelinderEtal2010}. Hinton \etal~\cite{hinton2015distilling} proposed the teacher-student framework for transferring such knowledge between DNNs being trained on the source and target tasks respectively. The high-level idea is to introduce an additional regularization for the DNN being trained on the target task, \ie, the student network, which allows learning the knowledge existing in the DNN that was pre-trained on the source task, \ie, the teacher network. While the framework was originally designed for knowledge transfer between DNNs on the same dataset, recent works \cite{yim2017gift, zagoruyko2016paying} started exploiting its potential for more general transfer learning tasks, \ie, when the source data and the target data are different.

Many knowledge transfer methods have been proposed with various intuitions. Hinton \etal~\citet{hinton2015distilling} and Ba and Caruana \citet{ba2014deep} propose to match the final layers of the teacher and the student network, as the outputs from the final layer of the teacher network provide more information than raw labels. Romero \etal~\citet{romero2014fitnets} proposes to match intermediate layers of the student network to the corresponding layers of the teacher network. Recent works \citep{belagiannis2018adversarial, ChenEtAl2018, huang2017like, yim2017gift, zagoruyko2016paying} relax the regularization of matching the entire layer by matching carefully designed features/statistics extracted from intermediate layers of the teacher and the student networks, \eg, attention maps \cite{zagoruyko2016paying} and maximum mean discrepancy \cite{huang2017like}.
%

Evidently, there is no commonly agreed theory behind knowledge transfer. This causes difficulty in understanding empirical results and in developing new methods in a more principled way. In this paper, we propose variational information distillation (VID) as an attempt towards this direction in which we formulate the knowledge transfer as maximization of the mutual information between the teacher and the student networks. This framework proposes an actionable objective for knowledge transfer and allows us to quantify the amount of information that is transferred from a teacher network to a student network. Since the mutual information is computationally intractable, we employ a variational information maximization \cite{agakov2004algorithm} scheme to maximize the variational lower bound instead. See Figure \ref{fig:diagram} for the conceptual diagram of the proposed knowledge transfer method. We further show that several existing knowledge transfer methods \cite{li2017learning, romero2014fitnets} can be derived as specific implementations of our framework by choosing different forms of the variational lower bound. We empirically validate the VID framework, which significantly outperforms existing methods. We observe the gap is especially large in the cases of small data and heterogeneous architectures.


In summary, the overall contributions of our paper are as follows:
\begin{itemize}
 \item We propose variational information distillation, a principled knowledge transfer framework through maximizing mutual information between two networks based on the variational information maximization technique.
 \item We demonstrate that VID generalizes several existing knowledge transfer methods. In addition, our implementation of the framework empirically outperforms state-of-the-art knowledge transfer methods on various knowledge transfer experiments, including knowledge transfer between (heterogeneous) DNNs on the same dataset or on different datasets.
\item Finally, we demonstrate that heterogeneous knowledge transfer between a convolutional neural networks (CNN) and a multilayer perceptrons (MLP) is possible on CIFAR-10.
Our method yields a student MLP that significantly outperforms the best-reported MLPs~\cite{lin2015far,UrbanEtAl2017} in the literature. \end{itemize}


\section{Variational information distillation (VID)}\label{sec:method}

In this section, we describe VID as a general framework for knowledge transfer in the teacher-student framework. Specifically, consider training a student neural network on a target task, given another teacher neural network pre-trained on a similar (or related) source task. Note that the source task and the target task could be the same, \textit{e.g.}, for model compression or knowledge distillation. The underlying assumption is that the layers in the teacher network have been trained to represent certain attributes of given inputs that exist in both the source task and the target task.
For a successful knowledge transfer, the student network must learn how to incorporate the knowledge of such attributes to its own learning.

From a perspective of information theory, knowledge transfer can be expressed as retaining high mutual information between the layers of the teacher and the student networks. More specifically, consider an input random variable $\bm{x}$ drawn from the target data distribution $p(\bm{x})$ and $K$ pairs of layers $\mathcal{R}=\{(\mathcal{T}^{(k)}, \mathcal{S}^{(k)})\}_{k=1}^{K}$, where each pair $(\mathcal{T}^{(k)}, \mathcal{S}^{(k)})$ is selected from the teacher network and the student network respectively. Feedforwarding the input $\bm{x}$ through the networks induces $K$ pairs of random variables $\{(\bm{t}^{(k)}, \bm{s}^{(k)})\}_{k=1}^{K}$ which indicate activations of the selected layers, \eg, $\bm{t}^{(k)} = \mathcal{T}^{(k)}(\bm{x})$. The mutual information between the pair of random variables $(\bm{t}, \bm{s})$ is defined by:
\begin{align}
I(\bm{t}; \bm{s}) &= H(\bm{t}) - H(\bm{t}|\bm{s}) \notag \\
&=-\mathbb{E}_{\bm{t}}[\log p(\bm{t})]
+\mathbb{E}_{\bm{t},\bm{s}}[\log p(\bm{t}|\bm{s})],
\end{align}
where the entropy $H(\bm{t})$ and the conditional entropy $H(\bm{t}|\bm{s})$ are derived from the joint distribution $p(\bm{t}, \bm{s})$. Empirically, the joint distribution $p(\bm{t}, \bm{s})$ is a result of aggregation over the layers with input $\bm{x}$ sampled from the input distribution $p(\bm{x})$. Intuitively, the definition of $I(\bm{t}; \bm{s})$ can be understood as a reduction in uncertainty in the knowledge of the teacher encoded in its layer $\bm{t}$ when the the student layer $\bm{s}$ is known.

\begin{figure*}[ht]
  \centering
  \begin{subfigure}{0.16\textwidth}
  \centering
  \includegraphics[width=0.95\textwidth]{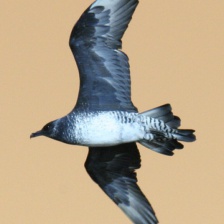}
  \vspace{.05in}\\
  \includegraphics[width=0.95\textwidth]{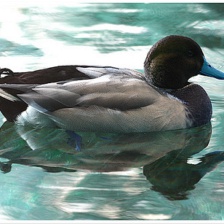}
  \vspace{.05in}\\
  \includegraphics[width=0.95\textwidth]{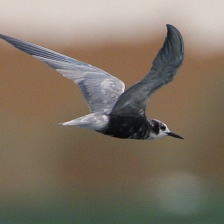}
  \caption{input}
  \label{fig:org}
  \end{subfigure}
  \hfill
  \begin{subfigure}{0.16\textwidth}
  \centering
  \includegraphics[width=0.95\textwidth]{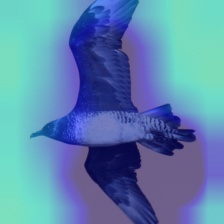}
  \vspace{.05in}\\
  \includegraphics[width=0.95\textwidth]{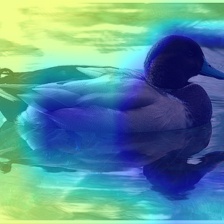}
  \vspace{.05in}\\
  \includegraphics[width=0.95\textwidth]{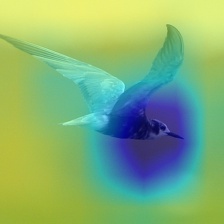}
  \caption{$0$-th epoch}
  \label{fig:log_prob1}
  \end{subfigure}
  \hfill
  \begin{subfigure}{0.16\textwidth}
  \centering
  \includegraphics[width=0.95\textwidth]{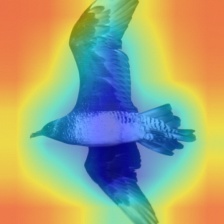}
  \vspace{.05in}\\
  \includegraphics[width=0.95\textwidth]{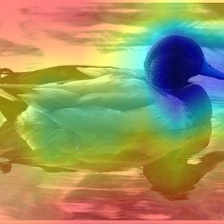}
  \vspace{.05in}\\
  \includegraphics[width=0.95\textwidth]{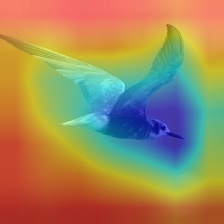}
  \caption{$40$-th epoch}
  \label{fig:log_prob2}
\end{subfigure}
  \hfill
  \begin{subfigure}{0.16\textwidth}
  \centering
  \includegraphics[width=0.95\textwidth]{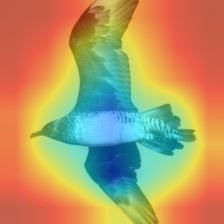}
  \vspace{.05in}\\
  \includegraphics[width=0.95\textwidth]{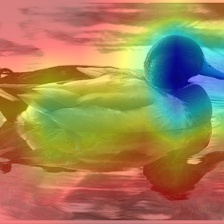}
  \vspace{.05in}\\
  \includegraphics[width=0.95\textwidth]{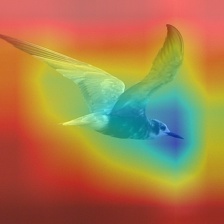}
  \caption{$160$-th epoch}
  \label{fig:log_prob3}
\end{subfigure}
  \hfill
  \begin{subfigure}{0.16\textwidth}
  \centering
  \includegraphics[width=0.95\textwidth]{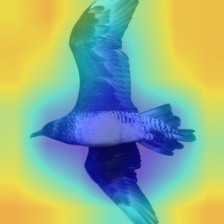}
  \vspace{.05in}\\
  \includegraphics[width=0.95\textwidth]{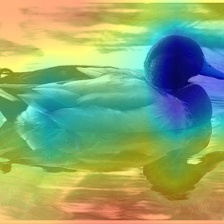}
  \vspace{.05in}\\
  \includegraphics[width=0.95\textwidth]{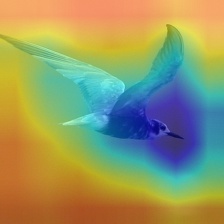}
  \caption{no transfer}
  \label{fig:vanilla}
  \end{subfigure}
  \hfill
  \begin{subfigure}{0.16\textwidth}
  \centering
  \includegraphics[width=0.95\textwidth]{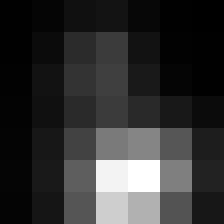}
  \vspace{.05in}\\
  \includegraphics[width=0.95\textwidth]{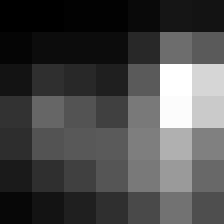}
  \vspace{.05in}\\
  \includegraphics[width=0.95\textwidth]{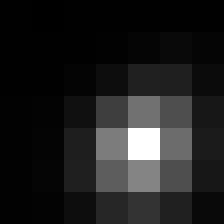}
  \caption{magnitude of $\bm{t}_{h,w}$}
  \label{fig:log_prob0}
  \end{subfigure}
\caption{
Plots for the heat maps corresponding to the variational distribution evaluated for spatial dimensions of the intermediate layer in the teacher network, \ie, $\log q(\bm{t}_{h,w}|\bm{s}) = \sum_{c} \log q(t_{c,h,w}|\bm{s})$. Each figure corresponds to (a) original input image, (b, c, d) log-likelihood $\log q(\bm{t}_{h,w}|\bm{s})$ that was normalized and interpolated to fit the spatial dimension of the input image (red pixels correspond to high probability), (d) log-likelihood of variational distribution optimized for the student network trained without any knowledge transfer applied and (f) magnitude of the layer $\bm{t}$ averaged for each spatial dimensions.
}
\label{fig:log_prob}
\end{figure*}

We now define the following loss function which aims to learn a student network for the target task while encouraging high mutual information with the teacher network:
\begin{equation}\label{eq:obj_exact}
  \mathcal{L} =
  \mathcal{L}_{\mathcal{S}}
  - \sum_{k=1}^{K}
  \lambda_{k}I({\bm{t}}^{(k)}, {\bm{s}}^{(k)}),
\end{equation}
where $\mathcal{L}_{\mathcal{S}}$ is the task-specific loss function for the target task and $\lambda_{k}>0$ is a hyper-parameter introduced for regularization of the mutual information in each layer. Equation \eqref{eq:obj_exact} needs to be minimized with respect to the parameters of the student network. However, the minimization is hard since exact computation of the mutual information is intractable. We instead propose a variational lower bound for each mutual information term $I(\bm{t}; \bm{s})$, in which we define a variational distribution $q(\bm{t}|\bm{s})$ that approximates $p(\bm{t}|\bm{s})$:
\begin{align}\label{eq:vim}
  &I(\bm{t}; \bm{s})=H(\bm{t})-H(\bm{t}|\bm{s}) \notag \\
  &= H(\bm{t})+ \mathbb{E}_{\bm{t}, \bm{s}}[\log p(\bm{t}|\bm{s})] \notag \\
  &= H(\bm{t})
  +\mathbb{E}_{\bm{t},\bm{s}}[\log q(\bm{t}|\bm{s})]
  +\mathbb{E}_{\bm{s}} [D_{\text{KL}}(p(\bm{t}|\bm{s}) || q(\bm{t}|\bm{s}))] \notag \\
  &\geq H(\bm{t})+\mathbb{E}_{\bm{t}, \bm{s}} [\log q(\bm{t}|\bm{s})],
\end{align}
where the expectations are over the distribution $p(\bm{t}, \bm{s})$ and the last inequality is due to the non-negativity of the Kullback-Leiber divergence $D_{\text{KL}}(\cdot)$. This technique is known as the \textit{variational information maximization} \citep{agakov2004algorithm}. Finally, we obtain VID by applying the variational information maximization to each mutual information term $I({\bm{t}}^{(k)}, {\bm{s}}^{(k)})$ in \eqref{eq:obj_exact}, leading to a minimization of the following loss function:
\begin{equation}\label{eq:obj}
  \widetilde{\mathcal{L}} =
  \mathcal{L}_{\mathcal{S}}
  -\sum_{k=1}^{K}\lambda_{k}
  \mathbb{E}_{\bm{t}^{(k)}, \bm{s}^{(k)}}
  [\log q(\bm{t}^{(k)}|\bm{s}^{(k)})].
\end{equation}
The objective $\widetilde{\mathcal{L}}$ is jointly minimized over the parameters of the student network and the variational distribution $q(\bm{t}|\bm{s})$. Note that the entropy term $H(\bm{t})$ has been removed from the equation \eqref{eq:vim} since it is constant with respect to the parameters to be optimized. Alternatively, one could interpret the objective \eqref{eq:obj} as jointly training the student network for the target task and maximization of the conditional likelihood to fit the activations of the selected layers from the teacher network. By doing so, the student network obtains the ``compressed'' knowledge required for recovering activations of the selected layers in the teacher network.

\subsection{Algorithm formulation}
We further specify our framework by choosing a form made for the variational distribution $q(\bm{t}|\bm{s})$. In general, we employ a Gaussian distribution with heteroscedastic mean $\bm{\mu}(\cdot)$ and homoscedastic variance $\bm{\sigma}$ as the variational distribution $q(\bm{t} | \bm{s})$, \ie, the mean $\bm{\mu}(\cdot)$ is a function of $\bm{s}$ and the standard deviation $\bm{\sigma}$ is not. Next, the parameterization of $\bm{\mu}(\cdot)$ and $\bm{\sigma}$ is further specified by the type of layer corresponding to $\bm{t}$. When $\bm{t}$ corresponds to intermediate layer of the teacher network with spatial dimensions indicating channel, height and width respectively, \ie, $\bm{t} \in \mathbb{R}^{C\times H \times W}$, our choice of variational distribution is expressed as follows:
\begin{align}
  \label{eq:conv}
  &-\log q(\bm{t}|\bm{s}) = -\sum_{c=1}^{C}\sum_{h=1}^{H}\sum_{w=1}^{W} \log q(t_{c, h, w}|\bm{s})\\
  &~~= \sum_{c=1}^{C}\sum_{h=1}^{H}\sum_{w=1}^{W}
  \log\sigma_{c}+\frac{(t_{c, h, w}-\mu_{c,h,w}(\bm{s}))^{2}}{2\sigma_{c}^{2}}+\text{constant} \notag,
\end{align}
where $t_{c,h,w}$ denote scalar components of $\bm{t}$ indexed by $(c,h,w)$. Further, $\mu_{c, h, w}$ represents the output of a single unit from the neural network $\bm{\mu}(\cdot)$ consisting of convolutional layers and the variance is ensured to be positive using the softplus function, \ie, $\sigma_{c}^{2} = \log (1+\exp(\alpha_{c}))+\epsilon$ where $\alpha_{c} \in \mathbb{R}$ being the parameter to be optimized and $\epsilon>0$ is minimum variance introduced for numerical stability. Typically, one can choose $\bm{s}$ from the student network with similar hierarchy and spatial dimension as $\bm{t}$. When spatial dimension of two layers are equal, $1\times1$ convolutional layers are typically used for efficient parameterization of $\bm{\mu}(\cdot)$. Otherwise, convolution or transposed convolution with larger kernel size could be used to match the spatial dimensions.

We additionally consider the case when the layer $\bm{t}=\mathcal{T}^{\text{(logit)}}(\bm{x})\in\mathbb{R}^{N}$ corresponds to the logit layer of the teacher network. Here, our choice of the variational distribution is expressed as follows:
\begin{align}
  \label{eq:fc}
  -\log q(\bm{t}|\bm{s}) &= -\sum_{n=1}^{N} \log q(t_{n}|\bm{s})\\
  &=\sum_{n=1}^{N}\log \sigma_{n}
  +\frac{(t_{n}-\mu_{n}(\bm{s}))^{2}}{2\sigma_{n}^{2}}
  +\text{constant}, \notag
\end{align}
where $t_{n}$ indicates the $n$-th entry of the vector $\bm{t}$, $\mu_{n}$ represents the output of a single unit of neural network $\bm{\mu}(\cdot)$ and $\sigma_{n}$ is, again, parameterized by softplus function to enforce positivity. For this case, the corresponding layer $\bm{s}$ in the student network is the penultimate layer $\mathcal{S}^{(\text{pen})}$ instead of the logit layer to match the hierarchy of two layers without being too restrictive on the output of the student network. Furthermore, we found that using a simple linear transformation for the parameterization of the mean function was sufficient in practice, \ie, $\bm{\mu}(\bm{s}) = \mathbf{W}\bm{s}$ for some weight matrix $\mathbf{W}$.

The aforementioned implementations turned out to perform satisfactorily during the experiments. We also considered using heteroscedastic variance $\bm{\sigma}(\cdot)$, but it gave unstable training with ignorable improvements. Other types of parameterizations such as a heavy-tailed distribution or the mixture density network \cite{bishop1994mixture} could be used to gain additional performance. We leave these ideas for future exploration.

See Figure \ref{fig:log_prob} for an illustration of the training VID using the implementation based on equation \eqref{eq:conv}. Here, we display the change in the evaluated log-likelihood of the variational distribution aggregated over channels, \ie, $\log q(\bm{t}_{h,w}|\bm{s}) = \sum_{c}\log q(t_{c,h,w}|\bm{s})$, given input $\bm{x}$ (Figure \ref{fig:org}) throughout the VID training process. One observes that the student network is trained gradually for the variational distribution to estimate the density of the intermediate layer from the teacher network (Figure \ref{fig:log_prob1}, \ref{fig:log_prob2} and \ref{fig:log_prob3}). As a comparison, we also optimize the variational distribution for the student network trained without knowledge transfer, (Figure \ref{fig:vanilla}). For this case, we observe that this particular instance of the variational distribution fails to obtain high log-likelihoods, indicating low mutual information between the teacher and the student networks. Interestingly, the parts that correspond to the background achieve higher magnitudes compared to that of the foreground in general. Our explanation is that the output of layers corresponding to the background that mostly corresponds to zero activations (Figure \ref{fig:log_prob0}) and contains less information, being a relatively easier target for maximizing the log-likelihood of the variational distribution.

\subsection{Connections to existing works}
\paragraph{The infomax principle.}
We first describe the relationship between our framework and the celebrated \emph{infomax principle} \citep{linsker1989application} applied to representation learning \citep{vincent2010stacked}, stating that ``good representation'' is likely to contain much information in the corresponding input. Especially, such a principle has been successfully applied to semi-supervised learning for neural networks by maximizing the mutual information between the input and output of the intermediate layer as a regularization to learning the target task, \eg, learning to reconstruct input based on autoencoders \citet{rasmus2015semi}. Our framework can be viewed similarly as an instance of semi-supervised learning with modification of the infomax principle: layers of the teacher network contain important information for the target task, and a good representation of the student network is likely to retain much of their information. One recovers the traditional semi-supervised learning infomax principle when we set $\bm{t}^{(k)}=\bm{x}$ in the equation \eqref{eq:obj_exact}.


\paragraph{Generalizing mean squared error matching.}
Next, we explain how existing knowledge transfer methods based on mean squared error matching can be seen as a specific instance of the proposed framework. In general, the methods will be induced from the equation \eqref{eq:obj} by making a specific choice of the layers $\mathcal{R}=\{(\mathcal{T}^{(k)}, \mathcal{S}^{(k)})\}_{k=1}^{K}$ for knowledge transfer and parameterization of heteroscedastic mean $\bm{\mu}(\cdot)$ in the variational distribution:
\begin{equation}
  \label{eq:mse}
  -\log q(\bm{t}| \bm{s}) = \sum_{n=1}^{N}
  \frac{(t_{n}-\mu_{n}(\bm{s}))^{2}}{2}+\text{constant}.
\end{equation}
Note that Equation \eqref{eq:mse} corresponds to a Gaussian distribution with unit variance over every dimension of the layer in the teacher network. Ba and Caruana~\citet{ba2014deep} showed that knowledge can be transferred between the teacher and the student networks that were designed for the same task, by matching the output of logit layers $\mathcal{T}^{(\text{logit})}, \mathcal{S}^{(\text{logit})}$ from the teacher and the student networks with respect to mean squared error. Such a formulation is induced from the equation \eqref{eq:mse} by letting $\mathcal{R}= \{(\mathcal{T}^{(\text{logit})}, \mathcal{S}^{(\text{logit})})\}$, and $\bm{\mu}(\bm{s})=\bm{s}$ in the equation \eqref{eq:mse}. This was later extended for knowledge transfer between the teacher and the student networks designed for different tasks by Li and Hoiem~\citet{li2017learning}, through adding an additional linear layer on top of the penultimate layer $\mathcal{S}^{(\text{pen})}$ in the student network to matching with logit layer $\mathcal{T}^{(\text{logit})}$ in the teacher network. This is induced similarly from the equation \eqref{eq:mse} by letting $\mathcal{R}=\{(\mathcal{T}^{(\text{logit})}, \mathcal{S}^{(\text{pen})})\}$, and $\bm{\mu}(\cdot)$ being a linear transformation, \ie, $\bm{\mu}(\bm{s})=\mathbf{W}\bm{s}$. Next, Romero \etal~\citet{romero2014fitnets} proposed a knowledge transfer loss for minimizing the mean squared error between intermediate layers from the teacher and the student networks, with additional convolutional layer introduced for adapting different dimension size between each pair of matched layers. This is recovered from the regularization term in the equation \eqref{eq:mse} by choosing layers for the knowledge transfer to be intermediate layers of the teacher and the student networks, and $\bm{\mu}(\cdot)$ being a linear transformation corresponding to a single $1 \times 1$ convolutional layer.

These methods are all similar to our implementation of the framework in that they all use Gaussian distribution as the variational distribution. However, our method differs in two key ways: (a) allowing the use of a more flexible nonlinear functions for heteroscedastic mean and (b) modeling different variances for each dimension in the variational distribution. This allows transferring mutual information in a more flexible manner without wasting model capacity. Especially, modeling unit variance for all dimensions of the layer $\bm{t}$ in the teacher network could be highly restrictive for the student network. To illustrate, the layer of the teacher network might include an activation $t_{n}$ that contains information irrelevant to the task of the student network, yet requires much capacity for regression of $\mu_{n}(\bm{s})$ to $t_{n}$. This would raise over-regularization issues, \ie, wasting the majority of the student network's capacity on trying to fit such a unit. Instead, modeling high homoscedastic variance $\sigma_{n}$ for such dimension make its contribution ignorable to the overall loss, allowing one to ``filter'' out such unit in an efficient way.

\paragraph{Comparison with feature matching.}
Besides the knowledge transfer methods based on mean squared error matching, several works \citep{ChenEtAl2018, huang2017like, yim2017gift, zagoruyko2016paying} have proposed indirectly matching the handcrafted features extracted from intermediate layers. More specifically, Zagoruyko and Komodakis~\citet{zagoruyko2016paying} proposed matching the ``attention maps'' generated from activations from the layers. Huang and Wang~\citep{huang2017like} later generalized the attention map to matching the maximum mean discrepancy of the activations. Yim \etal~\citep{yim2017gift} proposed matching the feature called the Flow of Solution Procedure (FSP) defined by the Gram matrix of layers adjacent in the same network. Chen \etal~\cite{ChenEtAl2018} considered matching the reconstructed input image from the intermediate layers of the teacher and the student networks. These methods could be seen as smartly avoiding the aforementioned over-regularization issue by filtering out information in the teacher network using expert knowledge. However, such methods potentially lead to suboptimal results when the feature extraction method is not apt for the particular knowledge transfer task and may discard important information from the layer of the teacher network in an irreversible way.

\section{Experiments}
%
We demonstrate the performance of the proposed knowledge transfer framework by comparing VID to state-of-the-art knowledge transfer methods on image classification. We apply VID to two different locations: (a) VID between intermediate layers of the teacher and the student network (VID-I) and (b) VID between the logit layer of the teacher network and the penultimate layer of the student network (VID-LP). For comparison, we consider the following knowledge transfer methods: the original knowledge distillation (KD) \citet{hinton2015distilling}, learning without forgetting (LwF) \citet{li2017learning}, hint based transfer (FitNet) \citep{zagoruyko2016paying}, activation-based attention transfer (AT) \citet{zagoruyko2016paying} and polynomial kernel-based neural selectivity transfer (NST) \citet{huang2017like}. Note that we consider FitNet as a regularization for training the student network \citep{zagoruyko2016paying} instead of a stage-wise training procedure as first proposed in \citep{romero2014fitnets}. We compare knowledge transfer methods for knowledge transfer between same and different datasets, which is commonly referred to as the knowledge distillation and transfer learning tasks respectively.

In all the experiments, we select the same pairs of intermediate layers for knowledge transfer based on VID-I, FitNet, AT and NST. Similarly, the same pairs of layers for knowledge transfer are used for LwF and VID-LP. All the hyper-parameters of all the methods are chosen according to the performance on a validation set, which is 20\% of the training set. We carefully pick the set of candidate values of hyper-parameters such that all the values proposed in the original works are included. The presented performances are the average of three repeated runs. More details about experiments are included in the supplementary material. The implementation of the algorithm will be made publicly available shortly. 

\subsection{Knowledge distillation}
\label{sec:distil}

We first compare knowledge transfer methods on the traditional knowledge distillation task, where a student network is trained on the same task as the teacher network. By distilling the knowledge from a large teacher network into a small student network, we can speed up the computation for prediction. We further investigate two problems for this task: whether we can benefit from knowledge transfer in the small data regime and how much performance we lose by reducing the size of the student network? Note that we do not evaluate the performance of VID-LP and LwF as they are designed for transfer learning. When applied, KD, VID-LP and LwF delivered similar performance.

\begin{table}[t]
	\small
		\centering
		\begin{tabular}{lcccc}
				\toprule
			$M$ & 5000 & 1000 & 500 & 100 \\
				\midrule
			Teacher & 94.26 & - & - & - \\
			Student & 90.72 & 84.67 & 79.63 & 58.84 \\
				\midrule
			KD & 91.27 & 86.11 & 82.23 & 64.24 \\
			FitNet & 90.64 & 84.78 & 80.73 & 68.90 \\
			AT & 91.60 & 87.26 & 84.94 & 73.40 \\
			NST & 91.16 & 86.55 & 82.61 & 64.53 \\
			VID-I & \textbf{91.85} & \textbf{89.73} & \textbf{88.09} & \textbf{81.59} \\
				\midrule
			KD + AT & 91.81 & 87.34 & 85.01 & 76.29 \\
			KD + VID-I & 91.7 & 88.59 & 86.53 & 78.48 \\
				\bottomrule
	\end{tabular}
	\caption{Experimental results (test accuracy) of knowledge distillation on the CIFAR-10 dataset from teacher network (WRN-40-2) to student network (WRN-16-1) with varying number of data points per class (denoted by $M$).}
	\label{tab:distil_nsc}
\end{table}

\paragraph{Reducing training data.}
Knowledge transfer can be a computationally expensive task. Given a pre-trained teacher network on the whole training data set, we explore the possibility of using a small portion of the training set for knowledge transfer. We demonstrate the effect of a reduced training set by applying knowledge distillation on CIFAR-10 \citep{krizhevsky2009learning} with four different sizes of training data. We employ wide residual networks (WRN) \citep{krizhevsky2009learning} for the teacher network (WRN-40-2) and the student network (WRN-16-1), where the teacher network is pre-trained on the whole training set of CIFAR-10. Knowledge distillation is applied to four different sizes of training set: 5000 (the full size), 1000, 500, 100 data points per class.

We compare VID-I with KD, FitNet, AT and NST. We also provide performances of the teacher network (Teacher) and the student network trained without any knowledge transfer (Student) as baselines. We choose four pairs of intermediate layers similarly to \citep{zagoruyko2016paying}, each of which is located at the end of a group of residual blocks. We implemented VID-I using three $1\times 1$ convolutional layers with hidden channel size as twice of the output channel size.
\toappendix{
Hyper-parameter $\lambda$ for weighting
the knowledge transfer losses corresponding to
VID-I-0, VID-I-1, VID-I-2, KD, FiNet, AT and NST are chosen from
$[0.1, 1]$, $[1, 10]$, $[10, 100]$,
$[1.6, 16]$, $[10, 100]$, $[100, 1000]$
and $[5, 50]$ respectively.
We further vary the scaling of
cross entropy term from $[0.1, 1]$.
When training on the full dataset,
we used stochastic gradient descent for
$200$ epochs with batch size of $128$ and weight decay of $0.0005$.
Initial learning rate of $0.1$ is decayed $0.2$ times
at $60, 120, 160$-th epoch.
Otherwise,
the numbers are appropriately scaled to have
similar number of updates for parameters.
}
The results are shown in Table \ref{tab:distil_nsc}. Our method, VID-I, outperforms other knowledge transfer methods consistently across all regimes. The performance gap increases as the size of dataset get smaller, \eg, VID-I only drops $10.26\%$ of accuracy even when $100$ data points per each class are provided to the student network. There is a $31.88\%$ drop without knowledge transfer and a $15.52\%$ drop for the best baseline, \ie, KD + AT.


\begin{table}[t]
	\small
		\centering
		\begin{tabular}{lcccc}
				\toprule
			($d$, $w$) & (40,2) & (16, 2) & (40, 1) & (16, 1) \\
				\midrule
			Teacher & 74.16 & - & - & - \\
			Student & 74.34 & 70.42 & 68.79 & 65.46 \\
				\midrule
			KD & 75.80 & 72.87 & 70.99 & 66.03 \\
			FitNet & 74.29 & 70.89 & 68.66 & 65.38 \\
			AT & 74.76 & 71.06 & 69.85 & 65.31 \\
			NST &  74.81 & 71.19 & 68.00 & 64.95 \\
			VID-I & 75.25 & 73.31 & 71.51 & 66.32 \\
				\midrule
			KD + AT & 75.86 & 73.13 & 71.4 & 67.07 \\
			KD + VID-I & \textbf{76.11} & \textbf{73.69} & \textbf{72.16} & \textbf{67.19}\\
				\bottomrule
	\end{tabular}
	\caption{Experimental results (test accuracy) of knowledge distillation on the CIFAR-100 dataset from the teacher network (WRN-40-2) to the student networks (WRN-$d$-$w$) with varying factor of depth $d$ and width $w$.}
	\label{tab:distil_struct}
\end{table}

\paragraph{Varying the size of the student network.}
The size of the student network gives a trade-off between the speed and the performance in knowledge transfer. We evaluate the performance of knowledge transfer methods on different sizes of the student network. The teacher network (WRN-40-2) is pre-trained on the whole training set of CIFAR-100. A student network with four choices of size, \ie, WRN-40-2, WRN-16-2, WRN-40-1, WRN-16-1, is trained on the whole training set of CIFAR-100. We compare our VID-I with KD, FitNet, AT and NST along with the Teacher and Student baselines. The choices of intermediate layers are the same as the previous experiment.

The results are shown in in Table \ref{tab:distil_nsc}. As also noticed by Furlanello \etal~\cite{furlanello2018born}, the student network with the same size as the teacher network outperforms the teacher network with all the knowledge transfer methods. One observes that VID-I consistently outperforms FitNet, AT and NST, which correspond to the same choice of layers for knowledge transfer. It also outperforms KD except for the case when the structure of the student network is identical to that of the teacher network, \ie, WRN-40-2, where two methods can be combined to yield the best performance.


\begin{table*}[t]
	\small
\centering
	\begin{subtable}{.48\textwidth}
		\centering
		\begin{tabular}{lcccc}
      \toprule
      $M$ & $\approx$80 & 50 & 25 & 10 \\
      \midrule
        Student  & 48.13 &37.69&27.01&14.25 \\
        fine-tuning & 70.97 & 66.04&58.13&47.91 \\
			\midrule
			LwF & 63.43 & 51.79 & 41.04 & 22.76 \\
			FitNet & 71.34 & 60.45 & 54.78 & 36.94 \\
			AT & 58.21 & 48.66 & 43.66 & 27.01 \\
			NST & 55.52 & 46.34 & 33.21 & 20.82 \\
			VID-LP & 67.91 & 58.51 & 47.09 & 31.94 \\
			VID-I & 71.34 & 63.66 & 60.07 & \textbf{50.97} \\
			\midrule
			LwF + FitNet & 70.97 & 60.37 & 54.48 & 38.73 \\
			VID-LP + VID-I & \textbf{71.87} & \textbf{65.75} & \textbf{61.79} & 50.37 \\
			\bottomrule
    \end{tabular}
	\caption{MIT-67, ResNet-34 to ResNet-18}
	\label{tab:transfer_mit_homo}
	\end{subtable}
	\hfill
	\begin{subtable}{.48\textwidth}
	\centering
	\begin{tabular}{lcccc}
		\toprule
			$M$ & $\approx$80 & 50 & 25 & 10 \\
		\midrule
			Student  & 53.58 & 43.96 & 29.70 & 15.97 \\
			fine-tuning &	65.97 & 58.51 & 51.72 & 39.63 \\
			\midrule
		LwF & 60.90 & 52.01 & 41.57 & 27.76 \\
		FitNet & 70.90 & 64.70 & 54.48 & 40.82 \\
		AT &	60.90 & 52.16 & 42.76 & 25.60 \\
		NST & 55.60 & 46.04 & 35.22 & 21.64 \\
		VID-LP & 68.88 & 61.64 & 50.22 & 39.25 \\
		VID-I & \textbf{72.01} & \textbf{67.01} & \textbf{59.33} & \textbf{45.90} \\
			\midrule
		LwF + FitNet & 70.52 & 64.10 & 54.63 & 40.15 \\
		VID-LP + VID-I & 71.72 & 66.49& 58.96 & 45.89 \\
		\bottomrule
	\end{tabular}
	\caption{MIT-67, ResNet-34 to VGG-9}
	\label{tab:transfer_mit_hetero}
\end{subtable}
\vspace{.1in} \\
\begin{subtable}{.48\textwidth}
\centering
\begin{tabular}{lcccc}
	\toprule
		$M$ & $\approx$29.95 & 20 & 10 & 5 \\
	\midrule
		Student  & 37.22 & 24.33 & 12.00 & 7.09\\
		fine-tuning &	76.69 & 71.00 & 59.25 & 44.07 \\
		\midrule
	LwF & 55.18 & 42.13 & 26.23 & 14.27 \\
	FitNet & 66.63 & 56.63 & 46.68 & 31.04 \\
	AT &	54.62 & 41.44 & 28.90 & 16.55 \\
	NST & 55.01 & 41.87 & 23.76 & 15.63 \\
	VID-LP & 65.59 & 54.12 & 39.20 & 27.86 \\
	VID-I & \textbf{73.25} & \textbf{67.20} & \textbf{56.86} & \textbf{46.21} \\
		\midrule
	LwF + FitNet & 68.69 & 58.81 & 48.86 & 31.30 \\
	VID-LP + VID-I & 69.71 & 63.94 & 52.87 & 41.12 \\
	\bottomrule
\end{tabular}
\caption{CUB-200-2011, ResNet-34 to ResNet-18}
\label{tab:transfer_cub_homo}
\end{subtable}
\hfill
\begin{subtable}{.48\textwidth}
\centering
\begin{tabular}{lcccc}
	\toprule
		$M$ & $\approx$29.95 & 20 & 10 & 5 \\
	\midrule
		Student & 44.59	& 32.10	& 15.69	& 9.66 \\
		fine-tuning & 60.96	& 51.86	& 46.88	& 39.98 \\
		\midrule
	LwF & 52.18 & 38.05 & 25.57 &	13.93 \\
	FitNet & 68.96 & 61.52 & 48.04 & 32.89 \\
	AT & 56.28 & 43.96 & 28.33 & 13.98 \\
	NST & 56.55 & 44.95 & 28.43 & 14.66 \\
	VID-LP & 66.82 & 55.94 & 38.10 & 30.47 \\
	VID-I & \textbf{71.51} & \textbf{65.69} & 53.29 & 38.09 \\
		\midrule
	LwF + FitNet & 70.56 & 62.44 & 47.36 & 30.52 \\
	VID-LP + VID-I & 70.00	& 65.14 & \textbf{53.78} & \textbf{38.76}\\
	\bottomrule
\end{tabular}
\caption{CUB-200-2011, ResNet-34 to VGG-9}
\label{tab:transfer_cub_hetero}
\end{subtable}
	\caption{Experimental results (test accuracy) of transfer learning from the teacher network (ResNet-34) to the student network (ResNet-18/VGG-9) for the MIT-67/CUB-200-2011 dataset with varying number of data points per class (denoted by $M$). We use $M\approx M_{\text{avg}}$ to denote the setting where the number of data points per class is non-uniform and $M_{\text{avg}}$ in average. Fine-tuning gives good results on transfer learning, but is not directly comparable as it is not a knowledge transfer method.}
	\label{tab:transfer}
\end{table*}

\subsection{Transfer learning}
\label{sec:transfer}

We evaluate knowledge transfer methods on transfer learning. The teacher network is a residual network (ResNet-34) \citep{he2016deep} pre-trained on the ImageNet dataset \citep{russakovsky2015imagenet}. We apply transfer learning to improve the performance of two separate image classification tasks. The first task is a fine-grained bird species classification based on the CUB-200-2011 dataset \citep{WelinderEtal2010}, which contains 11,788 images in total for 200 bird species. The second task is an indoor scene classification based on the MIT-67 dataset \citep{quattoni2009recognizing}, which contains 15,620 images for 67 classes of indoor scenes. For both tasks, there are a relatively few images per class, which can significantly benefit from knowledge transfer from the ImageNet classification task. To evaluate the performance at various levels of data scarcity, we subsample both datasets into three different sizes (50, 25, 10 per class for MIT-67 and 20, 10, 5 per class for CUB-200-2011) and compare the knowledge transfer methods.

%

We evaluate the knowledge transfer methods in two scenarios: a smaller student network of the same architecture (ResNet-18) and different architecture (VGG-9) \citep{simonyan2014very}. We compare our VID-I and VID-LP with LwF, FitNet, AT and NST. We evaluate the performance of the student network without transfer learning (Student) as a baseline. For the teacher and the student network with ResNet architecture, we choose the outputs of the third and fourth groups of residual blocks (from the input) as the intermediate layers for knowledge transfer. In the case of the VGG-9 student network, we choose the fourth and fifth max-pooling layers as the intermediate layers for knowledge transfer, which corresponds to the same spatial dimension as the intermediate layers selected from the teacher network. For applying VID-I to the ResNet-18 student network, we use two $1\times 1$ convolutional layers with the size of intermediate channels as half of the output channel size. When the student network is VGG-9, a single $1\times 1$ convolutional layer without non-linearity is used.

The results are shown in Table \ref{tab:transfer}. The knowledge transfer from ResNet-34 to VGG-9 gives very similar performance to the transfer from ResNet-34 to ResNet-18 for all the knowledge transfer methods. This shows that knowledge transfer methods are robust against small architecture changes. Our methods outperform other knowledge transfer methods in all regions of comparison. Both VID-I and VID-LP outperforms baselines that correspond to the same choice of layers for knowledge transfer. For the MIT-67 dataset, we observe that our algorithm outperforms even the finetuning method, which requires pre-training of the student network on the source task.

\toappendix{
Hyper-parameter $\lambda$ considered for
VID-LP, VID-I-0, VID-I-1, VID-I-2, LwF, FitNet, AT and NST are
$[10, 100]$,
$[10, 100]$,
$[10, 100]$,
$[10, 100]$,
$[1.6, 16]$,
$[10, 100]$,
$[100, 1000]$ and
$[5, 50]$ respectively
with scaling of the original
cross entropy term chosen from $[0.1, 1]$.
When training on the full dataset for
ResNet-34 and ResNet-18,
we use SGD for $250$ epochs with
batch size $128$ and weight decay of $0.0005$.
Initial learning rate of $0.05$ is
decayed by $0.2$ times at $150, 200$-th epoch.
For the case of VGG-9,
we use SGD for $250$ epochs
with batch size $12$ without weight decay.
Initial learning rate of $0.01$ is
decayed by $0.2$ times at $150$ and $200$-th epoch.
When training on subset of the dataset,
the numbers are appropriately scaled to
match the number of updates for parameters.}

\subsection{Knowledge transfer from CNN to MLP}
\label{sec:mlp}
\begin{table}[ht]
	\small
\centering
		\begin{tabular}{l|cccc}
      \toprule
       Network & MLP-4096 & MLP-2048 & MLP-1024\\
      \midrule
			Student & 70.60 & 70.78 & 70.90\\
			KD & 70.42 & 70.53 & 70.79\\
			FitNet & 76.02 & 74.08 &72.91\\
			VID-I & \textbf{85.18} & \textbf{83.47} & \textbf{78.57}\\
			\midrule
			Urban \etal~\cite{UrbanEtAl2017} & & 74.32 & \\
			Lin \etal~\cite{lin2015far}& & 78.62 & \\
			\bottomrule
    \end{tabular}
		\vspace{0.1in}
	\caption{Experimental result (test accuracy) of distillation on CIFAR-10 from the convolutional teacher network (WRN-40-2) to
	the fully connected student network (MLP-$h$) with
	varying size of hidden dimensions $h$.}
	\label{tab:conv_to_fc}
\end{table}

The transfer learning experiments show the robustness of the knowledge transfer method against small architecture changes. This leads to an interesting question: whether a knowledge transfer method can work between two completely different network architectures. A solution to this question can open a new direction of knowledge transfer and potentially offer solutions to many problems, \eg, speeding up prediction of recurrent neural networks (RNNs) by transferring knowledge from a RNN to a CNN, speeding up prediction of CNN on CPU or low-energy device by transferring knowledge from a CNN to a multi-layer perceptron (MLP). 

In this paper, we evaluate the performance of knowledge transfer from CNN to MLP on CIFAR-10. There is a well-known performance gap between CNN and MLP on CIFAR-10 \cite{lin2015far, UrbanEtAl2017}. The state-of-the-art performance on CIFAR-10 with MLP is 78.62\% with initialization from auto-encoders  \cite{lin2015far} and 74.32\% using knowledge distillation \cite{UrbanEtAl2017}. Urban \etal~\cite{UrbanEtAl2017} also trained a single convolutional layer achieving the performance of 84.6\% using knowledge distillation.

We apply the knowledge transfer methods in the knowledge distillation setting as mentioned in Section \ref{sec:distil}. We use a teacher network with convolutional layers (WRN-40-2) pre-trained on CIFAR-10. We use a MLP with five fully connected hidden layers as the student network, constructed by stacking one linear layer, three bottleneck linear layers and one linear layer in sequence. Each is followed by a non-linearity activation in between. Here, the bottleneck layer indicates a composition of two linear layers without non-linearity that is introduced to speed up learning by reducing the number of parameters. All the hidden layers have the same $h$ number of units and the bottleneck linear layer is composed of two linear layers with a size of $h\times\frac{h}{4}$ and $\frac{h}{4}\times h$.

%
%

The knowledge transfer between intermediate layers is defined between the outputs of four residual groups of the teacher network and the outputs of the first four fully connected layers of the student network. We compare VID-I with KD and FitNet since these knowledge transfer methods do not rely on spatial structures. For the same reason, AT and NST are not applicable to multilayer perceptrons. VID-I is implemented with multiple transposed convolutional layers without non-linearities. Specifically, the inputs for the variational distributions, \ie, the hidden layers of the MLP are treated as a tensor with $1\times 1$ spatial dimensions. Single transposed convolutional layer with a $4\times 4$ kernel, unit stride and zero padding is followed by multiple transposed convolutional layers with a $4\times 4$ kernel, two strides, and single padding to match the spatial dimension of the corresponding layer of the teacher network for knowledge transfer. More details on implementations of the student network and the auxiliary distribution are in the supplementary material.
\toappendix{
Hyper-parameter $\lambda$ for weighting the knowledge transfer
losses corresponding to VID-I, KD and AT were
chosen from $[10, 100]$, $[1.6, 16]$ and $[10, 100]$.
Scaling of cross entropy is chosen from $[0.1, 1]$.
For training, we used stochastic gradient descent (SGD) for
$700$ epochs with batch size of $128$ and
weight decay of $0.0005$.
Initial learning rate of $0.001$ was decayed by $0.2$ times
at $500$ and $600$-th epoch.
}

The results are shown in Table \ref{tab:conv_to_fc}. Both FitNet and VID-I improve the performance comparing the baseline of directly training the intermediate layers of the student network. VID-I significantly outperforms FitNet on MLPs with different sizes. Furthermore, MLP-4096 outperforms the the state-of-the-art performance with MLP reported by Lin \etal~\cite{lin2015far} (78.62\%) and Ba \etal~\cite{UrbanEtAl2017} (74.32\%) significantly. More importantly, our method bridges the performance gap between CNN (84.6\% using one convolutional layer \cite{UrbanEtAl2017}) and MLP shown in previous works.

\section{Conclusion}
In this work, we proposed the VID framework for effective knowledge transfer by maximizing the variational lower bound of the mutual information between two neural networks. The implementation of our algorithm is based on Gaussian observation models and is empirically shown to outperform other benchmarks in the distillation and transfer learning tasks. Using more flexible recognition models, \eg, \cite{kingma2016improved}, for accurate maximization of mutual information and alternative estimation of mutual information, \eg, \cite{belghazi2018mine}, are both ideas of future interest.

\newpage
{\small
\bibliographystyle{ieee}
\bibliography{reference}
}
\newpage
\appendix
\twocolumn[
  \begin{@twocolumnfalse}
    \begin{center}{\textbf{\LARGE Supplementary Material:}}
    \end{center}

    \begin{center}{\textbf{\Large Variational Information Distillation for Knowledge Transfer}}
    \end{center}
    \vspace{0.3in}
\end{@twocolumnfalse}
]
\section{Implementation details}
\subsection{Network architectures}
For the WRNs and ResNets used throughout the experiments,
we use the same architectures as originally described by
Zagoruyko \etal, \cite{zagoruyko2016wide}
and He \etal, \cite{he2016deep} respectively.
For the VGG-9 network used in transfer learning,
\ie, Section \ref{sec:transfer},
we make a slight modification from the VGG-11 network \cite{simonyan2014very}
without deviating from the VGG design philosophy.
It is conducted by first stacking eight $3\times 3$ convolutional layers with $64, 128, 256, 256, 512, 512, 512, 512$ channels in order with batch normalization and rectified linear unit (ReLU) after every convolutional layers. Furthermore, additional max-pooling layers are inserted after the $\{1,2,4,6,8\}$-th ReLUs.
Then the final max-pooling layer is followed by
global average pooling layer and a linear layer
leading up to the prediction of the labels.
For the MLP used in knowledge transfer
from CNN to MLP, \ie, Section \ref{sec:mlp},
we sequentially stack one linear layer, three bottleneck linear layers and one linear layer leading to the prediction of labels,
where dropout with drop rate of $0.2$, batch normalization
and ReLU was inserted between each layers.
Here, the bottleneck layer indicates a composition of two linear layers
without non-linearity that is introduced to speed up learning by
reducing the number of parameters.
All of the hidden layers have the same $h$ number of units and the
bottleneck linear layer is composed of two linear layers with a size of
$h\times\frac{h}{4}$ and $\frac{h}{4}\times h$.

\subsection{Parameterization of VID}
In the knowledge distillation experiments, \ie, Section \ref{sec:distil},
we parameterize the mean function $\bm{\mu}(\cdot)$ in equation \eqref{eq:conv}
for VID-I by three $1\times 1$ convolutional
layers with batch normalization and ReLU between
each layers. The hidden channel sizes were chosen to be
twice of the output channel size.
For the transfer learning experiments,
\ie, Section \ref{sec:transfer},
we first parameterize the mean function $\bm{\mu}(\cdot)$ in equation \eqref{eq:conv}
for VID-I by two $1\times 1$ convolutional layers with
batch normalization and ReLU between the layers.
For this case, the hidden channel sizes were
chosen to be half of the output channel size.
Furthermore, VID-LP
was parameterized as in equation \eqref{eq:fc}
with mean function $\bm{\mu}(\cdot)$ being a
single linear layer, \ie, a linear transformation.
Finally, we consider the knowledge transfer from CNN to MLP,
\ie, Section \ref{sec:mlp}, based on VID-I with equation \eqref{eq:conv}.
For this case, the mean function maps the one-dimensional input $\bm{s}$
from intermediate layer of the student network (MLP)
into a three-dimensional output $\bm{t}$
corresponding to intermediate layer of the teacher network (CNN), \ie,
$\bm{\mu}: \mathbb{R}^{N}\rightarrow \mathbb{R}^{C\times H \times W}$.
To this end, the input is first treated as a three-dimensional tensor
with with unit spatial dimensions, \ie,
$\bm{s} \in \mathbb{R}^{N\times 1 \times 1}$.
Then the input goes through a single transposed convolutional layer with
a $4\times 4$ kernel, unit stride and zero padding followed
by multiple transposed convolutional layers with a $4\times 4$ kernel,
two strides and unit padding.
The number of transposed convolutional layers were varied for
corresponding layer of the teacher network,
in order to match the spatial dimension.

\subsection{Loss function and training scheme}
In the experiments, the loss function for VID takes the following form:
\begin{equation}
\widehat{\mathcal{L}} =
\lambda_{1}\mathcal{L}_{\mathcal{S}}
- \sum_{k=1}^{K}
\frac{\lambda_{2}}{N_{k}}
\mathbb{E}_{\bm{t}^{(k)}, \bm{s}^{(k)}}
[\log q(\bm{t}^{(k)}|\bm{s}^{(k)})],
\end{equation}
where $\mathcal{L}_{\mathcal{S}}$ is the
task-specific loss function for the target task,
$\lambda_{1}, \lambda_{2} > 0$ are hyper-parameters
introduced for balancing between the
cross-entropy and the regularization terms,
and $N_{k}$ denotes the total number of dimensions for each
layer selected from the teacher network for knowledge transfer,
i.e., $\bm{t}^{(k)} \in \mathbb{R}^{N_{k}}$ or
$N_{k} = C_{k}H_{k}W_{k}$
when $\bm{t}^{(k)} \in \mathbb{R}^{C_{k}\times H_{k} \times W_{k}}$.
For all of the experiments and both VID-I and VID-LP,
we select $\lambda_{1}$ and $\lambda_{2}$
from $\{0.1, 1\}$ and $\{10, 100\}$ respectively,
based on the performance
evaluated on the validation set.
For other knowledge transfer methods,
we also choose the scaling of the cross-entropy term,
\ie, $\lambda_{1}>0$, from $\{0.1, 1\}$.
Furthermore, the corresponding regularization terms
are scaled by $\{1, 10\}, \{10, 100\}, \{100, 1000\}, \{5, 50\}$
for KD, FitNet, AT and NST respectively,
based on the performance on the validation set.
Additionally, KD was implemented with temperature scaling parameter set to $T=4$.

Finally, we describe the training scheme used for the experiments.
Due to unstable gradients in some cases,
we clipped the norm of gradients by $100$ throughout the experiments.
Additionally, the homoscedastic variance for the
variational distribution in equation \eqref{eq:conv} and \eqref{eq:fc} was
initialized with value of $5.0$.
In the knowledge distillation experiments, \ie, Section \ref{sec:distil},
when training on the full dataset,
we used stochastic gradient descent (SGD) for
$200$ epochs with batch size of $128$ and weight decay of $0.0005$.
Initial learning rate of $0.1$ is decayed $0.2$ times
at $\{60, 120, 160\}$-th epoch.
When training on subset of the dataset,
the numbers are appropriately scaled to have
similar number of updates for parameters.
In the transfer learning experiments, \ie, Section \ref{sec:transfer},
when training on the full dataset for
ResNet-34 and ResNet-18,
we use SGD for $250$ epochs with
batch size $128$ and weight decay of $0.0005$.
Initial learning rate of $0.05$ is
decayed by $0.2$ times at $\{150, 200\}$-th epoch.
For the case of VGG-9,
we use SGD for $250$ epochs
with batch size $12$ without weight decay.
Initial learning rate of $0.01$ is
decayed by $0.2$ times at $150$ and $200$-th epoch.
Again, the numbers are appropriately scaled to
match the number of updates for parameters
when training on subset of the full dataset.
In the knowledge transfer from CNN to MLP experiments,
\ie, Section \ref{sec:mlp}, we used SGD for
$700$ epochs with batch size of $128$ and weight decay of $0.0005$.
Initial learning rate of $0.001$ was decayed by $0.2$ times
at $500$ and $600$-th epoch.

\begin{figure}[t]
  \scriptsize
\begin{center}
\includegraphics[width=0.4\textwidth]{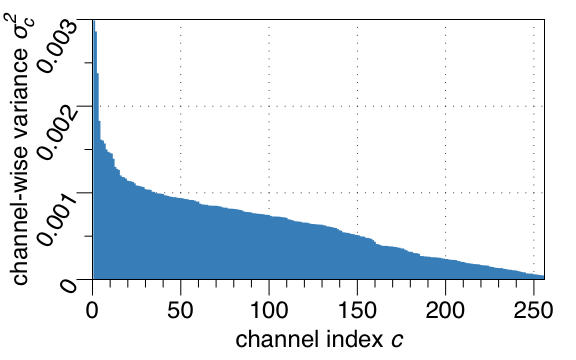}\\
\includegraphics[width=0.4\textwidth]{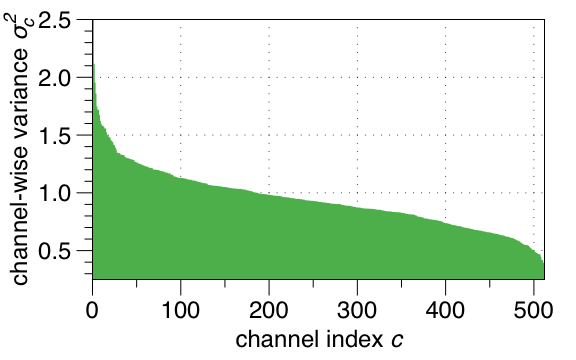}
\end{center}
   \caption{\it
   Channel-wise variance $\sigma_n^2 = \sigma_{c}^{2}$ (sorted) learned by VID-I in transfer learning
   from ResNet34 trained on ImageNet to ResNet18 trained on CUB-200,
   corresponding to the ends of third (top) and fourth (bottom)
   residual blocks respectively.}
\label{fig:variance}
\end{figure}

\begin{table}[t]
  \small
\centering
\resizebox{\linewidth}{!}{%
		\begin{tabular}{ccccccc}
      \toprule
       Student & KD & FitNet & AT & NST & VID-L & VID-I \\
      \midrule
			93.34 & 93.58 & 93.43 & 92.89 & 94.01 & 93.88 & \textbf{94.17}\\
			\bottomrule
    \end{tabular}}
	\caption{\it
    Experimental results (test accuracy)
    of transfer learning from grayscale-SVHN to MNIST with
    $200$ samples per class for LeNet-based architectures.}
	\label{tab:svhn}
\end{table}

\begin{table}[t]
  \small
\centering
\resizebox{\linewidth}{!}{%
		\begin{tabular}{ccccccc}
      \toprule
      Teacher & Student & KD & FitNet & ANC & VID-I \\
      \midrule
			92.36 / \underline{93.43} & 91.69 / \underline{91.42}
      & 91.12 & \underline{91.61} & \underline{91.92} &
      \textbf{92.15}\\
			\bottomrule
    \end{tabular}
  }
  \caption{\it
    Experimental results (validation accuracy) in comparison to
    Adversarial Network Compression (ANC),
    for knowledge distillation from ResNet-164 to ResNet-20 on CIFAR-10 dataset.
    Underlined numbers are results reported by Belagiannis \etal~\cite{belagiannis2018adversarial}.}
	\label{tab:adversarial}
\end{table}

\section{Additional Experiments}
\subsection{Visualization of learned parameters}
In order to examine the behavior of the learned variance parameters $\sigma_{n}$ in
VID, we plot its channel-wise value for different layers in Figure \ref{fig:variance}.
Here, one can observe that
the learned variance parameters $\sigma_{n}$ are diverse,
especially accross different layers.
Hence, modeling of homoscedastic variance is necessary for
obtaining a tighter lower bound of the mutual information
in the equation \eqref{eq:vim}.

\subsection{Transfer learning from SVHN to MNIST}
We also provide additional experimental results for transfer learning
from SVHN to MNIST in Table \ref{tab:svhn}.
To this end, the teacher network is trained on the
full SVHN dataset that was converted to grayscale,
then the student network is trained on MNIST with
$200$ data points per class.
We employ LeNet-like architectures for both networks.
Again, one observes that VID outperforms over
other methods.

\subsection{Comparison with adversarial network compression}
We additionally compare with the
recently proposed adversarial network compression \cite{belagiannis2018adversarial}.
by repeating the knowledge distillation experiment
on CIFAR-10 between ResNets presented by \cite{belagiannis2018adversarial}.
The corresponding results are reported in Table \ref{tab:adversarial}.
One observes that our methods outperforms the ANC with a small margin.



\subsection{Experimental results with standard deviation}
In Table \ref{tab:supp1}, \ref{tab:supp2}, \ref{tab:supp3}, \ref{tab:supp4}, 
\ref{tab:supp6} and \ref{tab:supp7}, we provide full experimental results
corresponding to the Table \ref{tab:distil_nsc}, \ref{tab:distil_struct},
\ref{tab:transfer_mit_homo}, \ref{tab:transfer_mit_hetero},
\ref{tab:transfer_cub_hetero} and \ref{tab:conv_to_fc} respectively with
additional standard deviations for the three repeated runs.

\subsection{Additional heat maps for VID training}
\begin{figure*}[ht]
  \centering
  \begin{subfigure}{0.16\textwidth}
  \centering
  \includegraphics[width=0.95\textwidth]{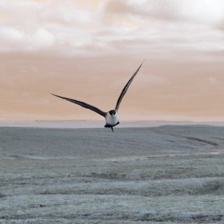}
  \vspace{.05in}\\
  \includegraphics[width=0.95\textwidth]{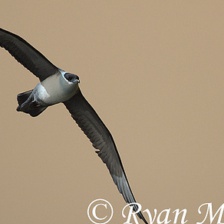}
  \vspace{.05in}\\
  \includegraphics[width=0.95\textwidth]{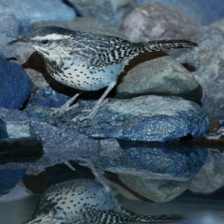}
  \vspace{.05in}\\
  \includegraphics[width=0.95\textwidth]{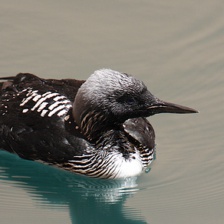}
  \vspace{.05in}\\
  \includegraphics[width=0.95\textwidth]{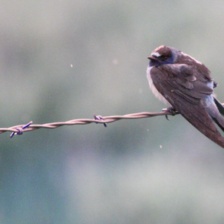}
  \vspace{.05in}\\
  \includegraphics[width=0.95\textwidth]{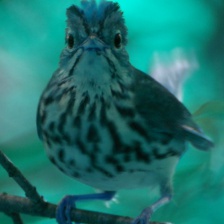}
  \vspace{.05in}\\
  \includegraphics[width=0.95\textwidth]{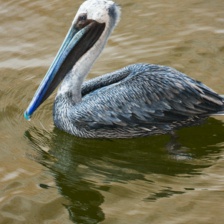}
  \caption{input}
  \end{subfigure}
  \hfill
  \begin{subfigure}{0.16\textwidth}
  \centering
  \includegraphics[width=0.95\textwidth]{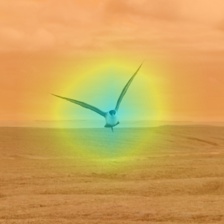}
  \vspace{.05in}\\
  \includegraphics[width=0.95\textwidth]{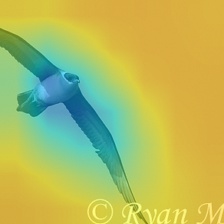}
  \vspace{.05in}\\
  \includegraphics[width=0.95\textwidth]{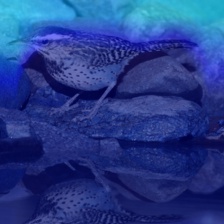}
  \vspace{.05in}\\
  \includegraphics[width=0.95\textwidth]{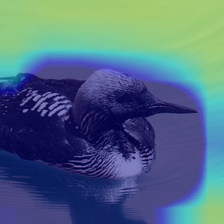}
  \vspace{.05in}\\
  \includegraphics[width=0.95\textwidth]{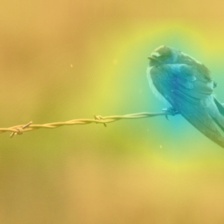}
  \vspace{.05in}\\
  \includegraphics[width=0.95\textwidth]{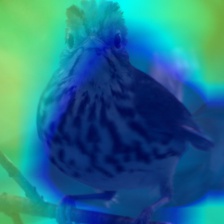}
  \vspace{.05in}\\
  \includegraphics[width=0.95\textwidth]{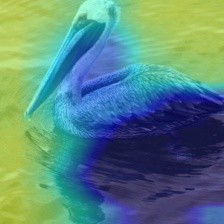}
  \caption{$0$-th epoch}
  \end{subfigure}
  \hfill
  \begin{subfigure}{0.16\textwidth}
  \centering
  \includegraphics[width=0.95\textwidth]{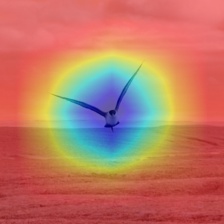}
  \vspace{.05in}\\
  \includegraphics[width=0.95\textwidth]{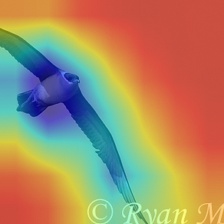}
  \vspace{.05in}\\
  \includegraphics[width=0.95\textwidth]{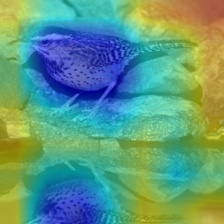}
  \vspace{.05in}\\
  \includegraphics[width=0.95\textwidth]{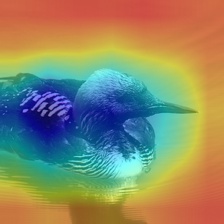}
  \vspace{.05in}\\
  \includegraphics[width=0.95\textwidth]{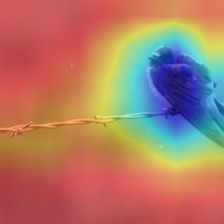}
  \vspace{.05in}\\
  \includegraphics[width=0.95\textwidth]{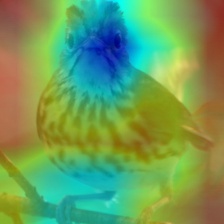}
  \vspace{.05in}\\
  \includegraphics[width=0.95\textwidth]{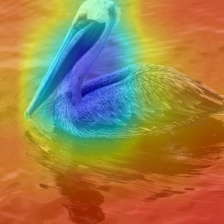}
  \caption{$40$-th epoch}
\end{subfigure}
  \hfill
  \begin{subfigure}{0.16\textwidth}
  \centering
  \includegraphics[width=0.95\textwidth]{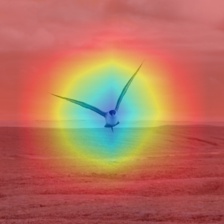}
  \vspace{.05in}\\
  \includegraphics[width=0.95\textwidth]{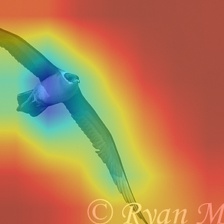}
  \vspace{.05in}\\
  \includegraphics[width=0.95\textwidth]{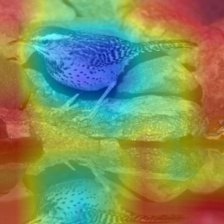}
  \vspace{.05in}\\
  \includegraphics[width=0.95\textwidth]{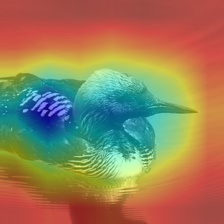}
  \vspace{.05in}\\
  \includegraphics[width=0.95\textwidth]{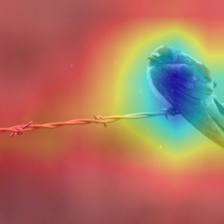}
  \vspace{.05in}\\
  \includegraphics[width=0.95\textwidth]{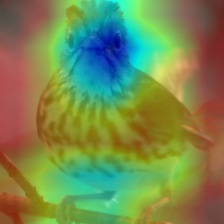}
  \vspace{.05in}\\
  \includegraphics[width=0.95\textwidth]{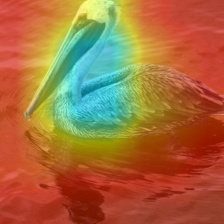}
  \caption{$160$-th epoch}
\end{subfigure}
  \hfill
  \begin{subfigure}{0.16\textwidth}
  \centering
  \includegraphics[width=0.95\textwidth]{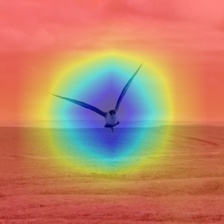}
  \vspace{.05in}\\
  \includegraphics[width=0.95\textwidth]{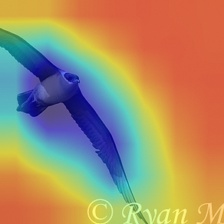}
  \vspace{.05in}\\
  \includegraphics[width=0.95\textwidth]{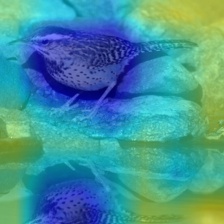}
  \vspace{.05in}\\
  \includegraphics[width=0.95\textwidth]{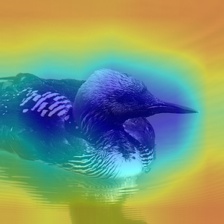}
  \vspace{.05in}\\
  \includegraphics[width=0.95\textwidth]{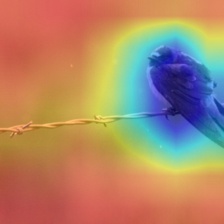}
  \vspace{.05in}\\
  \includegraphics[width=0.95\textwidth]{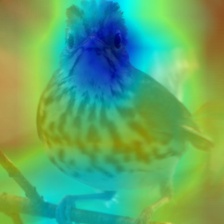}
  \vspace{.05in}\\
  \includegraphics[width=0.95\textwidth]{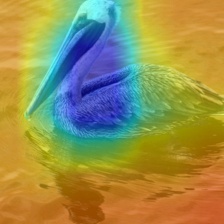}
  \caption{no transfer}
  \end{subfigure}
  \hfill
  \begin{subfigure}{0.16\textwidth}
  \centering
  \includegraphics[width=0.95\textwidth]{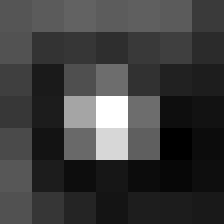}
  \vspace{.05in}\\
  \includegraphics[width=0.95\textwidth]{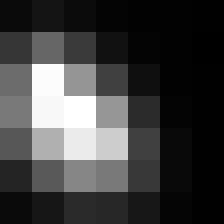}
  \vspace{.05in}\\
  \includegraphics[width=0.95\textwidth]{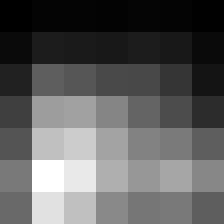}
  \vspace{.05in}\\
  \includegraphics[width=0.95\textwidth]{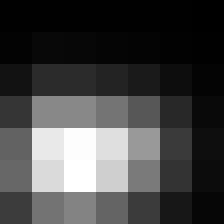}
  \vspace{.05in}\\
  \includegraphics[width=0.95\textwidth]{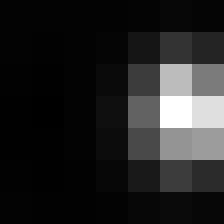}
  \vspace{.05in}\\
  \includegraphics[width=0.95\textwidth]{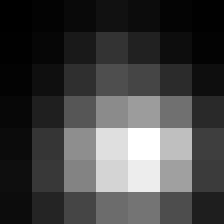}
  \vspace{.05in}\\
  \includegraphics[width=0.95\textwidth]{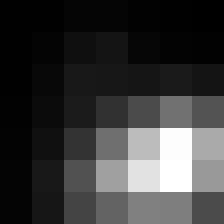}
  \caption{magnitude of $\bm{t}_{h,w}$}
  \end{subfigure}
\caption{Plots for the heat maps corresponding to the variational distribution evaluated for spatial dimensions of the intermediate layer in the teacher network, \ie, $\log q(\bm{t}_{h,w}|\bm{s}) = \sum_{c} \log q(t_{c,h,w}|\bm{s})$. Each figure corresponds to (a) original input image, (b, c, d) log-likelihood $\log q(\bm{t}_{h,w}|\bm{s})$ that was normalized and interpolated to fit the spatial dimension of the input image (red pixels correspond to high probability), (d) log-likelihood of variational distribution optimized for the student network trained without any knowledge transfer applied and (f) magnitude of the layer $\bm{t}$ averaged for each spatial dimensions.}
\label{fig:log_prob_supp}
\end{figure*}

In Figure \ref{fig:log_prob_supp},
we provide additional visualization results of the knowledge transfer based on VID that was plotted in the same way as in Figure \ref{fig:log_prob}.


\begin{table*}[tb]
	\small
		\centering
		\begin{tabular}{lcccc}
				\toprule
			$M$ & 5000 & 1000 & 500 & 100 \\
				\midrule
			Teacher & 94.36 \std{0.27} & - & - & - \\
			Student & 90.82 \std{0.17)} & 84.64 \std{0.05} & 79.64 \std{0.05}& 55.03 \std{6.59}\\
				\midrule
			KD & 91.66 \std{0.13} & 85.52 \std{0.02} & 81.48 \std{0.24} & 55.03 \std{0.05} \\
			FitNet & 90.79 \std{0.31} & 84.84 \std{0.35} & 80.82 \std{0.19} & 68.57 \std{0.84} \\
			AT & 91.54 \std{0.10} & 87.43 \std{0.35} & 84.78 \std{0.27} & 73.96 \std{0.96} \\
			NST & 91.11 \std{0.12} & 86.76 \std{0.37} & 82.68 \std{0.13} & 64.76 \std{0.45} \\
			VID-I & 91.94 \std{0.31} & \textbf{89.76} \std{0.07} & \textbf{88.33} \std{0.43} & \textbf{82.03} \std{1.13} \\
				\midrule
			KD + AT & 91.39 \std{0.26} & 87.11 \std{0.03} & 84.54 \std{0.01} & 75.11 \std{0.83} \\
			KD + VID-I & \textbf{92.31} \std{0.31} & 89.33 \std{0.21} & 87.34 \std{0.19} & 81.80 \std{0.01} \\
				\bottomrule
	\end{tabular}
	\caption{Experimental results (test accuracy) of knowledge distillation on the CIFAR-10 dataset from teacher network (WRN-40-2) to student network (WRN-16-1) with varying number of data points per class (denoted by $M$).}
	\label{tab:supp1}
\end{table*}

\begin{table*}[tb]
	\small
		\centering
		\begin{tabular}{lcccc}
				\toprule
			($d$, $w$) & (40,2) & (16, 2) & (40, 1) & (16, 1) \\
				\midrule
			Teacher & 74.16 \std{0.33} & - & - & - \\
			Student & 74.34 \std{0.46} & 70.42 \std{0.63} & 68.79 \std{0.19} & 65.46 \std{0.13} \\
				\midrule
			KD & 75.54 \std{0.25} & 72.94 \std{0.38} & 71.34 \std{0.19} & 66.97 \std{0.46}\\
			FitNet & 74.29 \std{0.17} & 70.89 \std{0.61} & 68.66 \std{0.27} & 65.38 \std{0.05} \\
			AT & 74.76 \std{0.36} & 71.06 \std{0.07} & 69.85 \std{0.51} & 65.31 \std{0.51} \\
			NST &  74.81 \std{0.19} & 71.19 \std{0.54} & 68.00 \std{0.20} & 64.95 \std{0.33} \\
			VID-I & 75.25 \std{0.37} & 73.31 \std{0.30} & 71.51 \std{0.15} & 66.32 \std{0.52} \\
				\midrule
			KD + AT & \textbf{75.90} \std{0.40} & 73.16 \std{0.15} & 71.48 \std{0.15} & 66.48 \std{0.67} \\
			KD + VID-I & \textbf{75.90} \std{0.26} & \textbf{73.50} \std{0.06} & \textbf{72.47} \std{0.21} & \textbf{66.91} \std{0.06}\\
				\bottomrule
	\end{tabular}
	\caption{Experimental results (test accuracy) of knowledge distillation on the CIFAR-100 dataset from the teacher network (WRN-40-2) to the student networks (WRN-$d$-$w$) with varying factor of depth $d$ and width $w$.}
	\label{tab:supp2}
\end{table*}

\begin{table*}[tb]
	\small
		\centering
		\begin{tabular}{lcccc}
      \toprule
      $M$ & $\approx$80 & 50 & 25 & 10 \\
      \midrule
        Student  & 48.78 \std{0.72} & 37.46 \std{0.88} & 25.52 \std{1.37} & 14.68 \std{0.41} \\
        Finetuned & 71.22 \std{0.85} & 65.30 \std{0.83} & 58.56 \std{0.55} & 48.86 \std{0.87}\\
			\midrule
			LwF & 61.34 \std{0.54} & 50.07 \std{0.22} & 38.76 \std{0.34} & 22.09 \std{0.58} \\
			FitNet & 70.37 \std{0.97} & 61.34 \std{0.94} & 54.60 \std{1.31} & 36.54 \std{0.34} \\
			AT & 57.99 \std{0.39} & 48.66 \std{0.67} & 42.51 \std{1.09} & 25.90 \std{1.29} \\
			NST & 56.79 \std{1.20} & 46.92 \std{0.80} & 34.38 \std{1.19} & 20.70 \std{0.22} \\
			VID-LP & 67.54 \std{0.42} & 59.18 \std{0.76} & 47.89 \std{0.75} & 31.22 \std{1.12} \\
			VID-I & \textbf{72.04} \std{0.62} & 66.42 \std{0.45} & 60.77 \std{0.91} & \textbf{50.60} \std{1.06} \\
			\midrule
			LwF + FitNet & 70.32 \std{0.69} & 61.19 \std{0.45} & 53.83 \std{0.91} & 36.67 \std{0.88} \\
			VID-LP + VID-I & 71.69 \std{0.37} & \textbf{66.87} \std{0.59}& \textbf{61.29} \std{0.04} & 49.65 \std{0.97}\\
			\bottomrule
    \end{tabular}
	\caption{Experimental results (test accuracy) of transfer learning from the teacher network (ResNet-34) to the student network (ResNet-18) for the MIT-67 dataset	with varying number of data points per class (denoted by $M$).}
	\label{tab:supp3}
\end{table*}

\begin{table*}[tb]
	\small
		\centering
		\begin{tabular}{lcccc}
			\toprule
				$M$ & $\approx$80 & 50 & 25 & 10 \\
			\midrule
				Student  & 54.13 \std{0.50} & 44.13 \std{0.30} & 29.05 \std{0.72} & 15.92 \std{0.67} \\
				Finetuned &	66.39 \std{0.41} & 58.51 \std{0.45} & 51.97 \std{0.31} & 39.93 \std{0.58} \\
				\midrule
			LwF & 58.18 \std{0.53} & 49.68 \std{2.09} & 38.08 \std{3.33} & 26.09 \std{1.08} \\
			FitNet & 71.00 \std{0.60} & 64.05 \std{0.63} & 55.30 \std{1.42} & 40.67 \std{0.13} \\
			AT &	60.57 \std{0.30} & 53.11 \std{0.83} & 42.64 \std{0.57} & 26.12 \std{0.52} \\
			NST & 55.40 \std{0.34}& 47.29 \std{1.23} & 34.03 \std{1.19} & 21.27 \std{0.71} \\
			VID-LP & 68.21 \std{0.59} & 61.77 \std{0.57} & 50.75 \std{0.49} & 39.23 \std{0.11} \\
			VID-I & \textbf{71.99} \std{0.19} & 66.62 \std{0.75} & \textbf{59.00} \std{0.38} & 46.24 \std{0.31}\\
				\midrule
			LwF + FitNet & 70.75 \std{0.47} & 64.38 \std{1.13} & 55.60 \std{0.13} & 41.34 \std{0.33}\\
			VID-LP + VID-I & 71.44 \std{1.21} & \textbf{66.67} \std{0.50} & 57.59 \std{0.23} & \textbf{46.42} \std{1.01} \\
			\bottomrule
		\end{tabular}
	\caption{Experimental results (test accuracy) of transfer learning from the teacher network (ResNet-34) to the student network (VGG-9) for the MIT-67 dataset	with varying number of data points per class (denoted by $M$).}
	\label{tab:supp4}
\end{table*}


\begin{table*}[tb]
	\small
		\centering
		\begin{tabular}{lcccc}
			\toprule
				$M$ & $\approx$29.95 & 20 & 10 & 5 \\
			\midrule
				Student & 44.59 \std{1.93}	& 32.10 \std{0.65}	& 15.69	\std{0.27} & 9.66 \std{0.22} \\
				Finetuned & 60.96	\std{1.88} & 51.86 \std{0.99} & 46.88 \std{0.92} & 39.98 \std{0.33} \\
				\midrule
			LwF & 52.54 \std{0.12} & 36.38 \std{0.14} & 22.79 \std{0.35} & 11.52 \std{0.15} \\
			FitNet & 68.96 \std{0.45} & 61.52 \std{0.80} & 48.04 \std{0.64} & 32.89 \std{1.95} \\
			AT & 56.28 \std{1.75} & 43.96 \std{0.80} & 28.33 \std{0.17} & 13.98 \std{1.01} \\
			NST & 56.55 \std{2.05} & 44.95 \std{0.36} & 28.43 \std{0.35} & 14.66 \std{2.48} \\
			VID-LP & 66.82 \std{0.41} & 55.94 \std{0.27} & 38.10 \std{0.83} & 30.47 \std{0.31} \\
			VID-I & \textbf{71.51} \std{1.48} & \textbf{65.69} \std{0.68} & \textbf{53.29} \std{1.20} & \textbf{38.09} \std{1.05}\\
				\midrule
				LwF + FitNet & 68.40 \std{0.50} & 61.40 \std{0.40} & 45.57 \std{0.04} & 28.41 \std{0.24}\\
				VID-LP + VID-I & 70.03 \std{0.05} & 63.46 \std{0.40} & 48.79 \std{0.04} & 32.35 \std{0.24} \\
			\bottomrule
		\end{tabular}
	\caption{Experimental results (test accuracy) of transfer learning from the teacher network (ResNet-34) to the student network (VGG-9) for the CUB-200 dataset	with varying number of data points per class (denoted by $M$).}
	\label{tab:supp6}
\end{table*}

\begin{table*}[tb]
	\small
		\centering
		\begin{tabular}{l|cccc}
      \toprule
       Network & MLP-4096 & MLP-2048 & MLP-1024\\
      \midrule
			Student & 70.60 \std{0.26} & 70.78 \std{0.45} & 70.90 \std{0.13}\\
			KD & 70.42 \std{0.26} & 70.53 \std{0.18} & 70.79 \std{0.35}\\
			FitNet & 76.02 \std{0.26} & 74.08 \std{0.18} &72.91 \std{0.35}\\
			VID-I & \textbf{85.18} \std{0.20} & \textbf{83.47} \std{0.29} & \textbf{78.57} \std{0.11} \\
			\bottomrule
    \end{tabular}
	\caption{Experimental result (test accuracy) of distillation on CIFAR-10 from the convolutional teacher network (WRN-40-2) to
	the fully connected student network (MLP-$h$) with
	varying size of hidden dimensions $h$.}
	\label{tab:supp7}
\end{table*}

\mbox{}
\clearpage
\newpage

\end{document}